%% file: PaGrKuCoLaneChangeSRNN2018.tex
\pdfoutput=1
\documentclass[letterpaper, 10 pt, conference]{ieeeconf}  

\IEEEoverridecommandlockouts                              

\usepackage{graphicx} 
\usepackage{epsfig} 
\usepackage{times} 
\usepackage{amsmath} 
\usepackage{amssymb}  
\usepackage{hyperref}

\DeclareMathOperator*{\argmax}{argmax}

\title{\LARGE \bf
Predicting Future Lane Changes of Other Highway Vehicles using RNN-based Deep Models
}

\author{Sajan Patel$^{1}$, Brent Griffin$^{1,2}$, Kristofer Kusano$^{3}$, and Jason J. Corso{$^{1,2}$}
\thanks{$^{1}$ {Robotics Institute, University of Michigan, Ann Arbor \newline \tt\footnotesize \{sajanptl, griffb, jjcorso\}@umich.edu}}%
\thanks{$^{2}$ {Electrical Engineering and Computer Science, University of Michigan, Ann Arbor \tt\footnotesize \{griffb, jjcorso\}@umich.edu }}%
\thanks{$^{3}$ {Toyota Motor North America Research and Development, Ann Arbor \tt\footnotesize{kris.kusano@toyota.com}}}
}

\begin{document}

\maketitle
\thispagestyle{empty}
\pagestyle{empty}

\begin{abstract}
In the event of sensor failure, autonomous vehicles need to safely 
    execute emergency maneuvers while avoiding other vehicles on the 
    road.  To accomplish this, the sensor-failed vehicle must predict 
    the future semantic behaviors of other drivers, such as lane changes, as 
    well as their future trajectories given a recent window of past sensor 
    observations.  We address the first issue of semantic behavior prediction 
    in this paper, which is a precursor to trajectory prediction, by introducing a framework that leverages the 
    power of recurrent neural networks (RNNs) and graphical models.  Our 
    goal is to predict the future categorical driving intent, for 
    lane changes, of neighboring vehicles up to three seconds into the future 
    given as little as a one-second window of past LIDAR, GPS, inertial, and 
    map data.
    
We collect real-world data containing over 20 hours of highway 
    driving using an autonomous Toyota vehicle. We propose a composite RNN model by adopting the methodology of Structural Recurrent Neural Networks (RNNs) to learn factor functions and take advantage of both the high-level structure of graphical models and the sequence modeling power of RNNs, which we expect to afford more transparent modeling and activity than 
    opaque, single RNN models.
    To demonstrate our approach, we validate our model using authentic 
	interstate highway driving to predict the future lane change maneuvers of 
	other vehicles neighboring our autonomous vehicle. We find that our composite Structural RNN outperforms baselines by as much as 12\% in
	balanced accuracy metrics.
\end{abstract}


\input{input/intro.tex}
\input{input/related.tex}
\input{input/data.tex}
\input{input/graph.tex}
\input{input/learn.tex}
\input{input/exp.tex}
\input{input/analysis.tex}

\section{Conclusion and Future Work} \label{sec:conclude}
We present a novel, lane-based SRNN for modeling and inferring the future lane
	change behavior expected to be made by neighboring exovehicles in highway
	settings.
We use SRNNs to map a transparent factor graph into a RNN architecture.
Our results and subsequent analysis shows detailed evidence that, first, our 
	model exhibits good performance for lane change prediction of exovehicles
	and, second, it has merit over different time horizon settings due to its
	lane-based structure.
While a few of the time horizon settings show mixed results, the extra 
	reliability and transparency afforded by our lane SRNN makes it a better 
	choice over the more opaque single LSTM and single-factor SRNN.
	
Future work in this problem space can focus on the specific failure modes of our
	lane SRNN.
Moreover, while this work specifically focuses on the problem of maneuver 
prediction on interstate highways, in which the key semantic maneuvers are 
limited to lane change, 
we note that our methods can be extended to a more diverse set of 
maneuvers present in city driving as well, such as turning at intersections.
Since much of the problem for maneuver anticipation for exovehicles 
	besides the given target vehicle based only on past LIDAR and inertial data 
	has been unexplored, we leave the extensions for city driving and more 
	diverse maneuvers for our future work.

\section*{Appendix}
\subsection{LSTM Equations}\label{app:lstm}
We provide the equations of a standard LSTM unit~\cite{LSTMref, jain-icra-2016} 
	for convenience.
Given in input sequence of feature vectors $x_t$, an initial hidden output
vector $h_0$, and an initial hidden context vector $c_0$, the follow operations
are carried out per time step:
\begin{align}
i_t &= \sigma(W_i x_t + U_i h_{t-1} + V_i c_{t-1} + b_i),\\
f_t &= \sigma(W_f x_t + U_f h_{t-1} + V_f c_{t-1} + b_f),\\
c_t &= f_t \odot c_{t-1} + i_t \odot \tanh(W_c x_t + U_c h_{t-1} + b_c),\\
o_t &= \sigma(W_o x_t + U_o h_{t-1} + V_o c_t + b_o),\\
h_t &= o_t \odot \tanh(c_t),
\end{align}
where $\odot$ is the element-wise product, $\sigma$ is the sigmoid 
function~\cite{murphy-ml-book}, and the various $W$, $U$, $V$ and $b$ matrices
and vectors are the weights and biases of the LSTM unit, respectively.
While these equations showcase the two recurrent functions modeled in a given
LSTM unit (hidden state and output), we note that in practice, we use a variant
of the LSTM that applies layer normalization before passing various quantities
through the activation functions~\cite{LeiBa2016layernorm}.

\subsection{Full Evaluation Results} \label{app:all-results}
We provide the full results for all evaluation metrics in Table 
\ref{tab:all-results} which are used in the analysis of our
methods.
\begin{table*}[h]
	\caption{Full Results for Time Horizon Analysis of Lane Change Prediction Methods}
	\label{tab:all-results}
	\centering
	\includegraphics[width=0.75\linewidth]{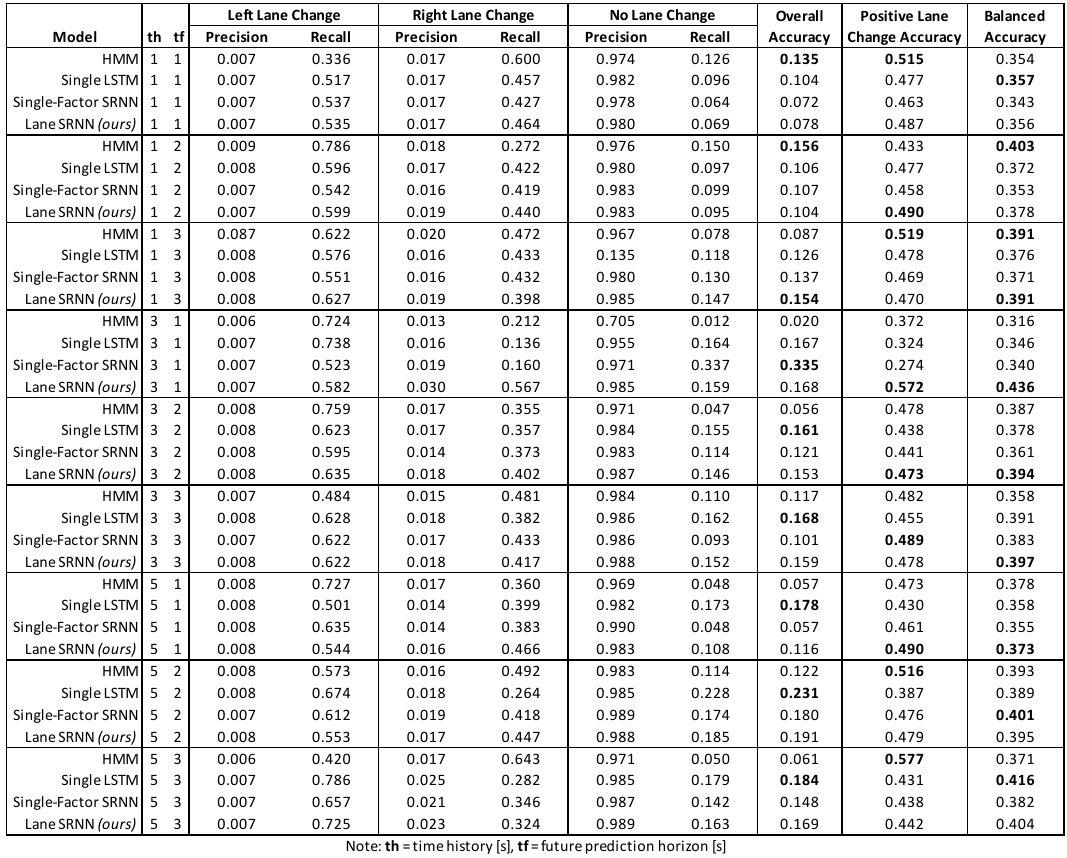}
\end{table*}

\section*{Acknowledgement}
This work was sponsored by Toyota Motor North America Research and Development
(TMNA R\&D). We would
like to acknowledge TMNA R\&D for providing the collected data set using the Toyota
Highway Teammate vehicle.  We also would like to acknowledge Richard Frazin for
his help in processing the data set.

\addtolength{\textheight}{-3cm}   

\bibliographystyle{IEEEtran}
\bibliography{./references.bib}

\end{document}

%% file: input/intro.tex
\section{Introduction}
Autonomous vehicles are equipped with many advanced sensors that allow them to 
perceive other vehicles, obstacles, and pedestrians in the environment.
Substantial work has been done in the areas of perception and reasoning for autonomous 
	vehicles and other forms of robots to allow these agents to make decisions
	based on their percepts~\cite{leonard2008perception, glaser-mtraj}.
  However, the majority of this past work make the tacit assumption that the 
  sensors are working reliably.  Hence, in these systems, the ability to make 
  autonomous decisions is lost under sensor failure.  In practice, such an 
  assumption is risky and not always a guarantee, especially in the case of 
  autonomous vehicles deployed in the real world amongst other human-driven 
  vehicles.  In the event of sensor failure in autonomous vehicles, only past 
  sensor readings are available for decision making.  These vehicles then need 
  to be able to plan and execute emergency maneuvers while safely avoiding 
  other moving obstacles on the road.
   
\begin{figure}[t]
  \centering
  \includegraphics[width=\linewidth]{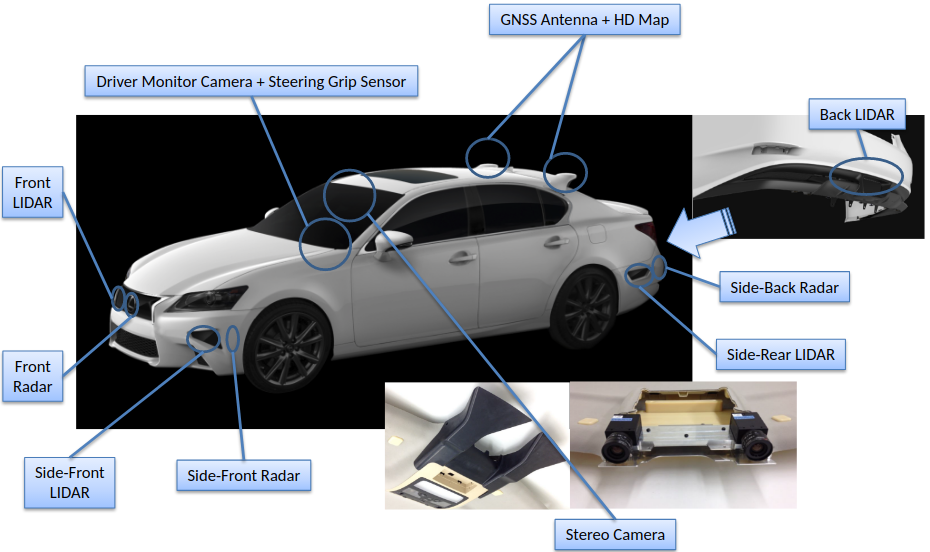}
  \caption{The Toyota autonomous vehicle used for data set collection and 
  experimentation. The sensor suite contains a multi-LIDAR system along with 
  GPS and inertial sensors. A stereo camera is also included and used for 
  visualizations, but it is not used in the data set for this work.}
  \label{fig:toyota-car}
\end{figure}

To optimally execute emergency maneuvers requires the knowledge of what other 
vehicles surrounding the \textit{blinded agent} are going to do in the near 
future, including predicting semantic maneuvers and 
	exovehicle\footnote{Throughout the paper, 
	we use the term exovehicles to refer to the vehicles in the vicinity of our autonomous vehicle, which we refer to as the egovehicle.} trajectories.  
In this work, we specifically address the first issue of predicting the 
maneuvers
of other vehicles up to three seconds into the future, such as performing left 
or right lane change maneuvers or staying in the same lane, while driving on 
the highway.
Since future exovehicle trajectories are dependent on these semantic behaviors, 
	tackling the behavior prediction problem becomes a necessary prerequisite
	that can simplify the trajectory prediction problem.
We approach this problem using as little as one-second and up to five-seconds 
	of past observations of neighboring vehicles
  based on LIDAR, GPS, inertial, and high-definition map data collected from an 
  autonomous Toyota vehicle (see Fig.~\ref{fig:toyota-car}).

There are two distinct categories of modeling choices for this problem of predicting semantic categories of exovehicle behavior: a classical probabilistic graphical modeling~\cite{koller2009probabilistic} approach and a contemporary deep neural network~\cite{GRUref, LSTMref} approach.  Both approaches are able to integrate various measurements into a common representation and to model their temporal evolution.

The classical approach of probabilistic graphical models~\cite{koller2009probabilistic},
such as factor graphs, spatiotemporal graphs, and dynamic Bayesian networks~\cite{MuPHD2002}, which bring graphical models into the 
sequential modeling space, is widely used in 
the robotics community for many reasons, including their interpretability and 
the high level structures, which can capture various relationships between 
features to modeling temporal sequences.  However, they require a 
parameterization of factor models that is structured by hand using 
domain-specific knowledge and optimized using various methods, including 
structural support vector machines and expectation maximization~\cite{jain-iccv-2015}, which, arguably, struggle to incorporate large-scale 
data well.  

On the other hand, recent advancements in temporal sequence modeling have come 
from the use of recurrent neural networks (RNNs)~\cite{GRUref, LSTMref}, which 
can be trained end-to-end for various tasks.  While methods that rely on deep 
learning lack the interpretability of factor graphs, these networks learn 
richer models than those currently employed in factor graphs.  Indeed, 
RNN-based methods have been applied to predicting future vehicle maneuvers but 
only in the context of making predictions for a single observed human 
driver~\cite{jain-icra-2016, jain-cvpr-2016}.

In this work, we want both the interpretability of factor graphs and the 
scalability of deep RNNs.  To that end, we bring RNN-based methods to the 
problem of predicting the future maneuvers of exovehicles within the vicinity of our own autonomous vehicle while traveling on a highway.  
We propose a composite RNN that leverages the recent work in Structural RNNs 
(SRNNs)~\cite{jain-cvpr-2016}.
Here, RNN units are connected in the form of factor graphs.
These networks employ the interpretable, high-level spatiotemporal structure of 
graphical models while using RNN units specifically to learn rich, nonlinear 
factor and node functions for factor graphs.  As with single RNN-based 
networks, SRNNs are trained end-to-end and can be unrolled over each 
step in the temporal sequence at inference time to make the predictions for the 
given task.  Following the methodology of SRNNs, we propose a novel 
lane-based graphical model which we then convert into a SRNN so
we can learn rich factor models for lane change prediction.

Our composite lane SRNN captures the spatiotemporal interactions between a 
given vehicle and its neighbors in the same and adjacent lanes.  To model 
lane-wise interactions, the graph includes a factor for the right, left, and 
same lanes that combines pose, velocity, and map-based lane information for the 
neighboring vehicles within the given lane.  
The model is unrolled over each time step of the sequence of past sensor 
observations to predict the future lane change maneuver class.
 
We provide an analysis on the efficacy of our lane-based SRNN in predicting 
  the future behavior of all tracked highway vehicles in alternative lanes, not 
  just forward and backward in the same lane, in the event of sensor 
  malfunction for varying future and past time horizons.
We train and evaluate our models using natural multi-lane 
interstate highway driving data obtained from an autonomous vehicle
  driving amongst other human drivers.
This data set is not augmented with simulated driving scenarios as in 
  Galceran et al., 2015~\cite{galceran2015rss} and is more extensive than other 
  highway datasets used in Jain et al., 2016~\cite{jain-icra-2016}.
Thus, the performance of our models on this data set constitutes the 
performance of our models on the actual autonomous robot for authentic highway 
driving.

%% file: input/related.tex
\section{Related Works}
\label{sec:related}
\subsection{Maneuver Anticipation}
Recent work in predicting driver maneuvers has primarily focused on
  the intent of the target vehicle's human driver; intent is based on tracking 
  the driver's face with an inward-facing camera along with features outside of 
  and in front of the vehicle using a camera, velocity sensor, and GPS 
  \cite{jain-cvpr-2016, jain-icra-2016, jain-iccv-2015}. 
The works of Jain et al., 2016~\cite{jain-cvpr-2016,jain-icra-2016} use various 
RNN-based architectures while Jain et al., 2015 uses graphical 
models~\cite{jain-iccv-2015}, both of which are essential to this work. 
However, rather than anticipating the behavior of our own vehicle, we address 
the problem of predicting the lane change maneuvers of neighboring vehicles in 
an interstate highway environment.
Moreover, by utilizing a multi-LIDAR system that provides all-around 
coverage, we predict future maneuvers for multiple vehicles in the surrounding 
neighborhood rather than only those detected and tracked in front of the 
data collection (ego) vehicle.

The method in Galceran et al., 2015~\cite{galceran2015rss} involves 
	anticipating the maneuvers of other vehicles and takes
  	a reinforcement learning approach to simulate multiple possible future 
  	maneuvers. 
While it chooses the ones that are most likely to occur; however, this approach 
is based on simulated approximations of limited highway driving scenarios.
Conversely, our method is trained and evaluated on data collected from natural 
freeway driving, which keeps our validation unaffected by simulation-based 
modeling errors.

\subsection{Graphical Models and Structural RNNs}
Graphical models are used in Jain et al., 2015~\cite{jain-iccv-2015} in the form of autoregressive
input-output Hidden Markov Models (HMMs) to model the temporal sequences that lead up 
to various maneuvers.
Similarly, HMMs are used in Schlechtriemen et. al, 2014~\cite{lc-feat-rank-ivs-2014}, using a similar neighborhood 
context for the target vehicle; however, this method relies on hand-tuned 
features computed from the tracked poses and map information of all of the vehicles rather than learning the 
factor models without restrictive assumptions on what features 
to extract.
The work presented in Jain et al., 2016~\cite{jain-cvpr-2016} bridges the gap 
between probabilistic 
graphical models and deep learning by introducing the Structural RNN, which 
exhibits better performance over graphical model counterparts through 
evaluations in many problem spaces, including maneuver anticipation for the 
target vehicle's human driver 
using facial tracking.
While our method follows the same methodology of transforming a graph 
into a Structural RNN, we propose a novel graph that takes into 
account lane-based spatiotemporal interactions between vehicles in the 
neighborhood
of the target vehicle to predict future lane change maneuvers.
We also evaluate the performance solely on natural freeway driving rather than 
city driving.

%% file: input/data.tex
\section{Problem Set-Up and Data}
\label{sec:data}
Given a recent history of sensor readings varying from one to five seconds, our
	goal is to predict the lane-changing behavior of exovehicles at prediction horizons varying from $1 \pm 0.5$ seconds to $3 \pm 0.5$ seconds.
Our prediction space is either left-lane change, right-lane change, or 
	no-lane change.
We collect a data set of highway driving using a Toyota sedan retrofitted with 
	the sensors of a typical automated vehicle.
In this paper, we will refer to this vehicle as the egovehicle, shown in 
	Figure~\ref{fig:toyota-car}.
The sensor suite includes 6 ibeo LUX 4L LIDARs mounted on all sides of the ego
	vehicle as well as an Applanix POS LV (version 5) high-accuracy GPS with
	Real-Time Kinetic (RTK) corrections.
Using the ibeo LUX Fusion System \cite{lux}, we detect the relative position 
	and orientation of neighboring vehicles up to approximately $120$ meters
	away. 

Using high-definition maps that include lane-level information (e.g. lane 
	widths, markings, curvature, GPS coordinates, etc.) along with GPS 
	measurements and the relative detections of neighboring vehicles, we 
	localize the ego vehicle and its neighbors on the map.
The GPS coordinates of the ego vehicle are projected into a world-fixed frame
  using the Mercator projection \cite{snyder1987map}, and the relative poses of 
  other vehicles are also mapped into this frame to determine absolute poses.
Similarly, the velocities and yaw rates of all vehicles are determined.
Along with vehicle poses, the maps allow us to determine the lane in which the 
	ego and neighboring vehicles are traveling in.
Over 20 hours of this data are collected at 12.5 Hz on multi-lane highways in 
	Southeast Michigan and Southern California, giving us roughly 1 million 
	samples for behavior prediction.

Given these off-the-shelf methods for detecting other vehicles and localizing
	them to the map, we focus on developing a framework that uses pose, 
	velocity, and lane information to predict future lane changes.
Specifically, we represent the $i^{th}$ vehicle at each time step $t$ with the 
	following state vector:
  	\begin{eqnarray}
  		v^i_t = [{P_x}^i_t, {P_y}^i_t, \psi^i_t, \dot {P_x}^i_t, \dot {P_y}^i_t, \dot \psi^i_t, n_l, n_r]^T,
  		\label{eq:stateVector}
  	\end{eqnarray}
	where $P_x$ and $P_y$ are the absolute world-fixed frame positions in 
	meters and $\dot P_x$ and $\dot P_y$ are their velocities, $\psi$ is the
	heading angle of the vehicle in radians with $\dot \psi$ as the yaw rate
	in radians/second, $n_l$ is the number of lanes to the left of the vehicle,
	and $n_r$ is the number of lanes to the right of the vehicle (both in the direction of travel).
We represent the sequence of historical states over time for each vehicle as
	\begin{eqnarray}
	V^i_t = [v^i_{t_h}, \dots, v^i_{t}],
	\label{eq:historicVector}
	\end{eqnarray}
	where $t_h$ is the maximum number of historical time steps included.
For each vehicle at each time step, there are three possible lane change 
	maneuvers that can occur $t_f \pm 7$ time steps into the 
	future\footnote{We use 7 steps to represent 0.5 seconds. Since the data 
	frequency is 12.5 Hz, we round all fractional steps up to the nearest 
	integer.}
	--left-lane change, right-lane change, and no-lane change.
We denote this set of possible maneuvers as $\mathcal{M} = \{left, right, no\}$.
These are determined by examining the change in lane identifiers provided 
	in the map between times $t_f - 7$ and $t_f + 7$.
We represent these labels as one-hot vectors $y^i_{t_f}$ in 
	\eqref{eq:stateVector} and annotate each vehicle state $V^i_t$ in
	\eqref{eq:historicVector} with its future lane change label vector.

%% file: input/graph.tex
\section{Lane Change Prediction Model}
  \begin{figure}[t]
      \centering
      \includegraphics[width=0.9\linewidth]{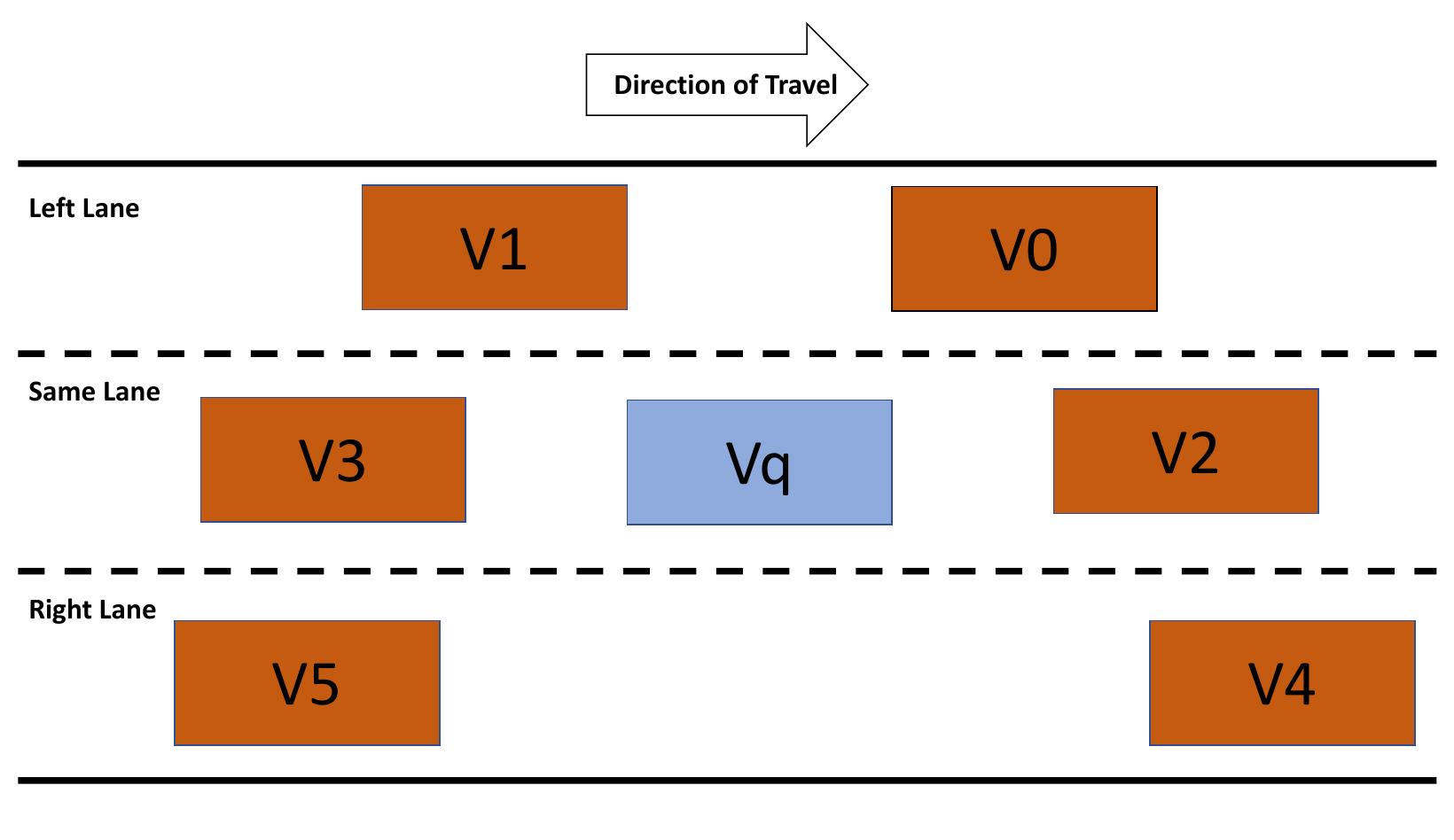}
      \caption{An example six-car (three-lane) context of neighboring
      vehicles surrounding the target exovehicle.
      We chose a six-car neighborhood since it is the minimal context required 
      for representing the lane-based interactions between the observed 
      vehicles. 
      All of these neighboring vehicles are not present at each time step in 
      natural freeway driving; thus, indicator variables are included 
      as a part of the augmented vehicle states used as inputs to our model.}
      \label{fig:context}
   \end{figure}
In this section, we discuss our proposed composite lane SRNN model, 
which allows us to transparently model that problem using factor graphs 
and, at the same time, compose the factors together into an RNN-based model.
For a given exovehicle $v^q$ traveling on a multi-lane highway, we model the 
	future lane change probability as a function of the vehicle's previous 
	states as well as the previous states of its neighboring vehicles.
We use a six-vehicle neighborhood, shown in Fig.~\ref{fig:context}, that 
  contains the vehicles ahead and behind the target vehicle in the left and 
  right lanes and the vehicles directly ahead and behind the target vehicle in 
  the same lane.
According to this convention, for every target vehicle $v^q$, the neighbors
	ahead and behind it in the left lane are $v^0$ and $v^1$, the neighbors 
	ahead and behind it in the same lane are $v^2$ and $v^3$, and the neighbors 
	ahead and behind it in the right lane are $v^4$ and $v^5$.

Since $v^q$ may be at the right- or left-most lane or other vehicles may be 
	out of sensor range during natural freeway driving, it is not guaranteed 
	that each of these neighboring positions is actually occupied by a vehicle.
For this reason, we only include a neighborhood of six vehicles, which provides 
	a minimalist representation of the target vehicle's context.
Accordingly, we augment the state of each neighboring vehicle $v^j$ from each 
	time step $t_h$ to $t$ with an indicator variable of 1 for when it is 
	present in the target vehicle's neighborhood and 0 for when it is not.
We use these augmented neighbor vehicle states as well as the target vehicle's
	state as the inputs to our lane change prediction model.

\subsection{Graphical Models for Lane-Based Maneuver Prediction}
\label{sec:graph}
Given the three-lane structure of the target vehicle's neighborhood, we design  
  	a factor graph that represents the probability of the future lane 
  	change label of the target vehicle using edges that represent the 
  	interaction between vehicles in each lane (left, same, right) with the 
  	target vehicle.
We make the assumption that given the observed state of the target 
	vehicle at a given time step, 
	the states of vehicles in a given lane are conditionally independent of vehicles in other lanes at that time step.
Hence, we model factors for vehicles in the left, right, and same lanes 
	separately.
Furthermore, we assume that the target vehicle behavior is conditioned on the
	states of the vehicles within the three-lane context.
The random variables are the target vehicle behavior labels $Y$ and the future
	vehicle states; however, only the future behavior label is of interest.
We start with the joint distribution over the target vehicle's behavior labels 
	and the states of all vehicles within the context from times $t_h$ to $t$, 
  	and we use our assumptions to model the probability of the label $Y_{t_f}$ 
  	taking on value $m \in \mathcal{M}$ as follows:
\begin{align}
 P&(Y_{t_f}=m, y_t, \dots, y_{t-t_h}, V^q_t, V^0_t, \dots, V^5_t)\label{eq:time-step-eqn}\\
&= \sum_{y_{t_f} \in \mathcal{M} / m}{P(y_{t_f}, y_t,\dots, y_{t-t_h}, V^q_t, V^0_t, \dots, V^5_t)}\nonumber\\
\begin{split}
&= \sum_{y_{t_f} \in \mathcal{M} / m} P(v^q_{t_h}, v^0_{t_h}, \dots, v^5_{t_h}) \, \times \nonumber \\ 
& \quad \quad \prod^{t}_{k=t_h}\Bigg(P(y_{k+1} | v^q_k, v^0_k, \dots, v^5_k)\\
 & \quad  \quad \quad \times \, P( v^q_k, v^0_k, \dots, v^5_k | v^q_{k-1}, v^0_{k-1}, \dots, v^5_{k-1})\Bigg)
\end{split} \nonumber
\end{align}
We further factorize the distributions in \eqref{eq:time-step-eqn} based on our 
	assumption of conditional independence between lanes and Markovian temporal 
	dynamics:
\begin{align}
P&(y_{k+1} | v^q_k, v^0_k, \dots, v^5_k) \label{eq:spatiotemporal-factors} \\
&= P(y_{k+1} | v^q_k, v^0_k, v^1_k) P(y_{k+1} | v^q_k, v^2_k, v^3_k) \, \times \nonumber \\
&\quad \quad P(y_{k+1} | v^q_k, v^0_k, v^4_k, v^5_k)\nonumber  \\
&= \phi_{l}(y_{k+1}, v^q_k, v^0_k, v^1_k) \phi_{s}(y_{k+1}, v^q_k, v^2_k, v^3_k) \, \times \nonumber \\
& \quad \phi_{r}(y_{k+1}, v^q_k, v^4_k, v^5_k)\nonumber
\end{align}
\begin{align}
P&(v^q_k, v^0_k, \dots, v^5_k | v^q_{k-1}, v^0_{k-1}, \dots, v^5_{k-1}) \label{eq:temporal-factors}\\
&= P(v^q_k, v^0_k, v^1_k | v^q_{k-1}, v^0_{k-1}, v^1_{k-1}) \nonumber \\
& \quad \times \, P(v^q_k, v^2_k, v^3_k | v^q_{k-1}, v^2_{k-1}, v^3_{k-1}) \nonumber\\
& \quad \times \, P(v^q_k, v^4_k, v^5_k | v^q_{k-1}, v^4_{k-1}, v^5_{k-1})\nonumber\\
&=  \gamma_{l}(v^q_k, v^0_k, v^1_k) \gamma_{s}(v^q_k, v^2_k, v^3_k) \gamma_{r}(v^q_k, v^4_k, v^5_k) \nonumber
\end{align}
	where each function $\phi(\cdot)$ and $\gamma(\cdot)$ is a parameterization 
	of the spatiotemporal and temporal factor functions, respectively, and 
	where $l$, $r$, $s$ denote the left, right, and same lanes. 
These functions can take on various forms, include exponential models in the 
	case of the spatiotemporal factors and Gaussians in the case of the 
	temporal models \cite{koller2009probabilistic}.
By parameterizing each of the three lane factors using the states of each of 
	the neighboring cars in the lane along with the target vehicle, we allow 
  	the model to take into account spatiotemporal interactions between each of 
  	the vehicles used in each factor.
Hence, from the vehicle states for a given lane, we can accommodate the 
	possibility of using relative distance and velocity features between
	vehicles.
The final lane change prediction is given by
\vspace{-2ex}
\begin{equation}
y^q_{t_f}=\argmax_{m \in \mathcal{M}}P(Y_{t_f}=m,y_t,...,y_{t-t_h},V^q_t,V^0_t,...,V^5_t).
\end{equation}
\vspace{-2ex}

%% file: input/learn.tex
\subsection{Learning Factor Functions using Structural RNNs}\label{sec:learn}
Factor functions are typically parameterized by hand to incorporate hand-tuned 
	features with simple weights, which limits the modeling power of standard 
	factor graphs~\cite{jain-cvpr-2016, jain-icra-2016, nowozin2011structured}.
Following the approach of Jain et al., 2016~\cite{jain-cvpr-2016}, we preserve 
	the transparency of the graphical model and, yet, leverage the power of 
	RNNs by converting it into a Structural RNN (SRNN) trained to classify the 
	lane change label.
The SRNN composites factors, captured as network snippets, into a larger RNN.
Specifically, random variable and factor nodes within graphical models are 
	represented using their own RNN units (which we will call \textit{nodeRNNs}
	and \textit{factorRNNs}, respectively).
This allows us to use the sequence modeling power of RNNs together with the 
	structure provided by our spatiotemporal factor graph.

To convert our graph into a Structural RNN, we use LSTM units to represent each 
	of the three lane interaction factors (each $\phi(\cdot)$ in 
	\eqref{eq:spatiotemporal-factors}) as \textit{factorRNNs}.
While standard Structural RNNs use different LSTM units for each spatiotemporal 
	and temporal factor as in Jain et al., 2016~\cite{jain-cvpr-2016}, 
  	we note that a single LSTM unit can jointly model both the spatiotemporal 
  	factors along with temporal factors (each $\gamma(\cdot)$ in 
  	\eqref{eq:temporal-factors}) for a given lane.
LSTM units have two recurrent functions within them--one for computing the 
	output and one for computing the context vector at each time step given 
	the input features and previous outputs and states~\cite{LSTMref}.
We provide extra details about the LSTM unit in Appendix A.
This allows us to model the spatiotemporal interaction factors with the 
	recurrent output function since the outputs of those are directly used in 
	the prediction of the future lane change label.
Similarly, we use the recurrent context vector function within each LSTM to 
	model temporal factors as a function of the input vehicle states.
By using a single LSTM to model both factors for each lane, we reduce the 
	complexity of our model and benefit from being able to train it with 
	a smaller dataset.

Since our graph has a single random variable node representing the future lane 
	change maneuver, our Structural RNN has one \textit{nodeRNN} to combine the 
	outputs of each lane's \textit{factorRNN}.
During the forward pass of the Structural RNN, each lane's vehicle state at 
	each time step are passed through their respective \textit{factorRNNs}.
The outputs of the three \textit{factorRNNs} are then concatenated passed
	through the \textit{nodeRNN}.
The following equations detail the computation performed at each time step, 
	noting that $LSTM$ represents the LSTM model (model details are provided 
	in Appendix A), $h^i_t$ and $c^l_k$ are the hidden outputs and context 
	vectors, respectively, of the $i^{th}$ LSTM unit at time step 
	$k \in [t_h, t]$, and that the scripts $l, r, s,$ and $n$ mean left lane, 
	right lane, same lane, and node, respectively:
\begin{align}
(h^l_k, c^l_k) &= LSTM_l([v^0_k; v^1_k; v^q_k], h^l_{k-1}, c^l_{k-1}),\\
(h^s_k, c^s_k) &= LSTM_s([v^2_k; v^3_k; v^q_k], h^s_{k-1}, c^s_{k-1}),\\
(h^r_k, c^r_k) &= LSTM_r([v^4_k; v^5_k; v^q_k], h^r_{k-1}, c^r_{k-1}),\\
(h^n_k, c^n_k) &= LSTM_n([h^l_k; h^s_k; h^r_k], h^n_{k-1}, c^n_{k-1}),
\end{align}
	where all $h^i_0$ and $c^i_0$ are zero initialized before each forward 
	pass through the network.
After unrolling all time steps of vehicle states through the Structural RNN, 
	we take the \textit{nodeRNN's} output of the last time step $h^n_t$ and 
	pass it through a softmax layer to obtain the final lane change prediction 
	for the target vehicle as follows:
\begin{align}
y^q_{t_f} &= \text{\texttt{softmax}}(W h^n_t + b), \label{eq:fc2-softmax}
\end{align}
	where $W$ and $b$ are the weights and bias of the fully connected layer 
	that transforms the output into the $|\mathcal{M}|$ log-probabilities fed 
	into the softmax function~\cite{bishop2006book, murphy-ml-book}.

At inference time, we use the final time step's output of the Structural RNN
	as the future lane change prediction; however, during end-to-end training, 
	we follow the approach of Jain et al., 2016~\cite{jain-icra-2016} and apply 
	a time-based, exponentially weighted softmax cross-entropy loss function to 
	each time step's output, whereby outputs of the network early on are 
	weighted less while outputs toward the end of the input sequence are
	weighted more.
This encourages the model to predict the label at all time steps while 	 
	penalizing early outputs less since they are only based on the early portion
	of the RNN input sequence.
Doing so leads to better recurrent outputs being used to build up to the final
	lane change prediction.
		
  \begin{figure}[t]
      \centering
      \includegraphics[width=0.75\linewidth]{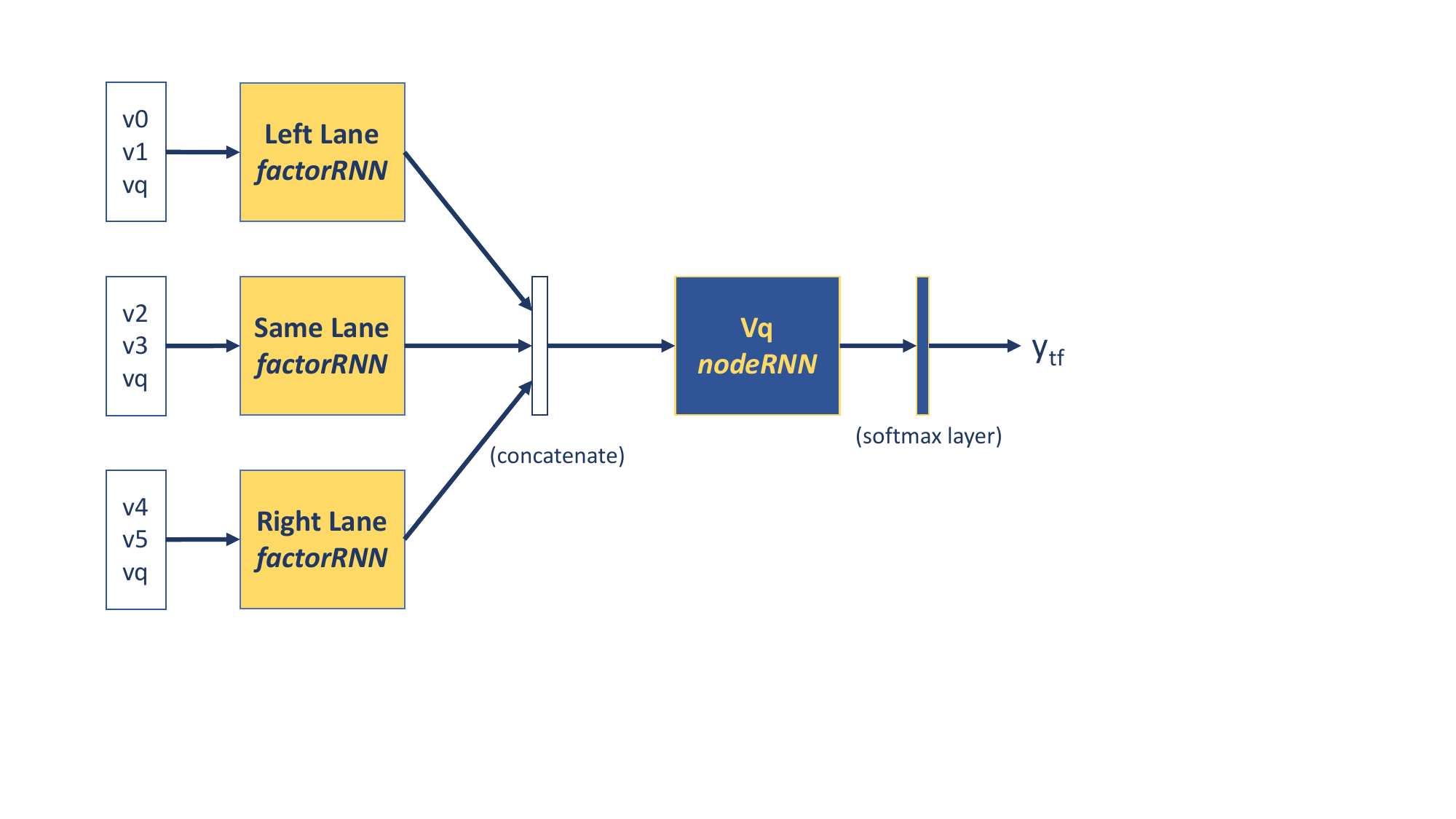}
      \caption{Our composite RNN model created using our lane-based factor 
          graph from Section~\ref{sec:graph}. At each $k$, the states for each 
      neighboring car for the given lane and the target car are concatenated 
      and fed as input through the given lane's \textit{factorRNN}. Each \textit{factorRNN} output is concatenated and fed through the \textit{nodeRNN}. Once the network is 
      unrolled across all the past $t_h$ observations, the last \textit{nodeRNN} output 
      is fed through one fully connected layer to get the final lane change 
      maneuver classification.}
      \label{fig:SRNN}
   \end{figure}

\subsection{Implementation Details}
Our proposed lane-based SRNN model is implemented using LSTMs 
	with layer normalization~\cite{LeiBa2016layernorm}.
All LSTM units within the SRNN have a hidden state size of 128.
The fully connected layer has an input size of 128 and output size of 
	$|\mathcal{M}| = 3$. 
During training, we use 50\% weight dropout in the LSTM units as implemented in 
	Semeniuta et al., 2016~\cite{semeniuta2016recurrentdropout}. 
We optimize the exponentially weighted softmax cross-entropy loss using the 
	ADAM optimizer~\cite{ADAM} with a learning rate of $10^{-4}$.
In addition, we center and scale the vehicle state inputs to have zero mean and 
	unit variance as a part of our preprocessing step.
All of our software is implemented in Python using 
Tensorflow~\cite{tensorflow2015-whitepaper}.

%% file: input/exp.tex
\section{Experiments}\label{sec:exp}
\subsection{Setup}
We use the data set collected by our autonomous vehicle (Sec. \ref{sec:data}) 
	for evaluation.
For a given time history and future prediction horizon, we sample all 
	the target vehicles tracked long enough to satisfy these time requirements 
	and split up the data set into a training and evaluation set that contains
	60\% and 40\% of the entire data, respectively.
In natural freeway driving scenarios, no-lane change events outnumber the 
	number of left-lane and right-lane change events; thus, we manually 
	balance each training set to contain equal numbers of left-lane, right-lane, and no-lane change samples.
We use the authentic, unbalanced evaluation set to test our models.

Each sample is pre-processed to center the initial target vehicle position and 
	orientation at the origin of a fixed reference frame.
This is done using applying a 2D rotation by the target vehicle's initial yaw 
	and a translation by its initial position coordinate to all vehicle 
	positions (for target and neighbors).
The initial target vehicle yaw is subtracted for all vehicles as well, and
	all velocities are rotated accordingly.
Following this step, we center the training data and scale it to have zero mean 
	and unit norm before passing the data as input to the lane change prediction
	model.

To evaluate the performance of our methods for various time horizons, we 
	train our lane-based SRNN and baseline models (Sec. \ref{ssec:baselines}) 
	from scratch for each setting of time history $t_h$ and future prediction 
	horizon $t_f$.
In our experiments, we choose 1, 3, and 5 seconds for 
	$t_h$\footnote{These correspond to 13, 38, and 63 time steps, 
		respectively when accounting for the data frequency of 12.5 Hz.}
	 and 1, 2, and 3
	seconds for $t_f$\footnote{Similarly, these correspond to 13, 25, and 38 
		time steps, respectively.}.
 
\subsection{Baseline Methods} \label{ssec:baselines}
We compare our lane-based SRNN to three types of baseline models--classical
	Hidden Markov models (HMMs), single LSTM models, and a simpler, 
	single-factor SRNN.
Since our method comes about from a temporal graphical model, we first
	compare it against the classical approach using Hidden Markov models (HMM) 
	\cite{murphy-ml-book}.
Each of the three behaviors is modeled using its own HMM with multivariate 
	Gaussian emission probabilities.
All the vehicle states (for neighbors and the target vehicle) are concatenated
	together to create a single observation vector per time step.
The HMMs are then trained in an unsupervised manner on training data specific 
	to their lane change class.
At inference time, the forward passes of the three HMMs are applied to the 
	input sequence in parallel to produce class probabilities, and the
	class with the highest (normalized) probability is chosen as the final 
	prediction.
We choose the number of latent states for each maneuver's HMM to maximize the 
	overall f1 score (harmonic mean of precision and recall) across the three 
	manuevers in a grid search evaluated on 20\% of the training data withheld for a validation set.
	
Since our method is composed of multiple LSTM units, we compare it against 
	a prediction model that uses only one LSTM unit.
This can also be viewed as having only a \textit{nodeRNN} present.	
To further evaluate the effect of our novel three-lane structure used in the 
	SRNN, we compare our method against a simpler, single-factor SRNN where we 
	only use one \textit{factorRNN} instead of three.
The output of this single \textit{factorRNN} is fed directly as input to the
	\textit{nodeRNN}.
This is akin to using a stacked LSTM for lane change prediction.

For both LSTM baselines, we use the same type of LSTM units with hidden state
	sizes of 128 and layer normalization as we do with our lane SRNN method.
The outputs at each time step of both models are also passed through the
	same size 128x3 fully connected layer and the softmax function to produce 
	the lane change probabilities.
At inference time, only the last time step's output is used to make a future 
	lane change prediction.
As with the HMM model, we concatenate all the vehicle states together per time
	step to feed as input to the model.
We train each LSTM baseline end-to-end using the exponentially weighted softmax
	loss function applied to all time steps, 50\% recurrent dropout, and 
	the same learning rate of $10^{-4}$ as with our lane SRNN.

\subsection{Evaluation Metrics} \label{Sec:Res}
We evaluate each model using precision, recall, and accuracy for predicting
	left, right, and no lane change behaviors for our proposed model and 
	baselines evaluated on authentic highway driving.
We compute the number of true positives $tp_m$, false positives $fp_m$, and 
	false negatives $fn_m$ for each maneuver $m \in \mathcal{M}$.
We use these to compute the precision $Pr_m$ and recall $Re_m$ values as 
	$Pr_m~=~tp_m~/~(tp_m~+~fp_m)$ and $Re_m~=~p_m~/~(tp_m~+~fn_m)$, 
	respectively.
We also calculate the overall prediction accuracy of the models, although it is 
	not very useful since the vast majority of the cases are no-lane change.	
To overcome this limitation, we calculate two other summary measures--balanced
	accuracy and positive lane-change accuracy. 
The balanced accuracy is a class averaged accuracy over the three cases, 
	equally weighting the left-lane, right-lane and no-lane change accuracies.
The positive lane-change accuracy is the accuracy for the subset of the 
	evaluation data with the no-lane change samples completely discarded.

%% file: input/analysis.tex
\section{Analysis} \label{sec:analysis}
We focus our analysis on the summary measures involving the various
	accuracy metrics; however, we provide the full results for each evaluation 
	metric (Sec \ref{Sec:Res}) in Table \ref{tab:all-results} in 
	Appendix B.
	\begin{table}
		\caption{Average of Accuracies Across All All Time Horizons}
		\label{tab:avg-accuracies}
		\begin{tabular}{r  c  c  c}
			\hline
			\rule{0pt}{2ex}	\bf Model & \bf Avg Acc & \bf Avg PLC Acc & \bf Avg Bal Acc\\
			\hline
			\rule{0pt}{2ex}HMM 				& 0.090	& 0.485	& 0.372\\
			Single LSTM 		& \bf 0.158 & 0.433	& 0.376\\
			Single-Factor SRNN	& 0.140 & 0.441	& 0.365\\
			\hline
			\rule{0pt}{2ex}Lane SRNN \textit{(ours)} 	& \textit{0.144} & \bf 0.487 & \bf 0.392\\
			\hline
		\end{tabular}
		\\\rule{0pt}{3ex}
		{Note: Acc refers to overall accuracy; PLC Acc refers to positive lane-change accuracy; Bal Acc refers to balanced accuracy.}
	\end{table}
\subsection{Holistic Performance}
To analyze how the models perform as a whole, we average the accuracy metrics
	provided in Table \ref{tab:all-results} and present them in Table 
	\ref{tab:avg-accuracies}.
We see that for average positive lane-change and balanced accuracies, our lane 
	SRNN outperforms all baselines.
The single LSTM does have a better average overall accuracy; however, this comes
	at the expense of missing the predictions of future lane changes while 
	predicting no-lane change behavior slightly better than the other models.
Even in this case, our lane SRNN is the second-best performing method, showing
	that it still has benefits over the single-factor SRNN and the classical 
	HMM baseline.
In general, all models are affected by false positives where no-lane change
	samples are incorrectly predicted as left or right lane changes.
This is a result of the skewed class representation in authentic driving and 
	leads to relatively small overall accuracies on the evaluation data.

\subsection{Comparison of RNN-based Methods Against HMMs}
\begin{figure}
	\centering
	\includegraphics[width=\linewidth]{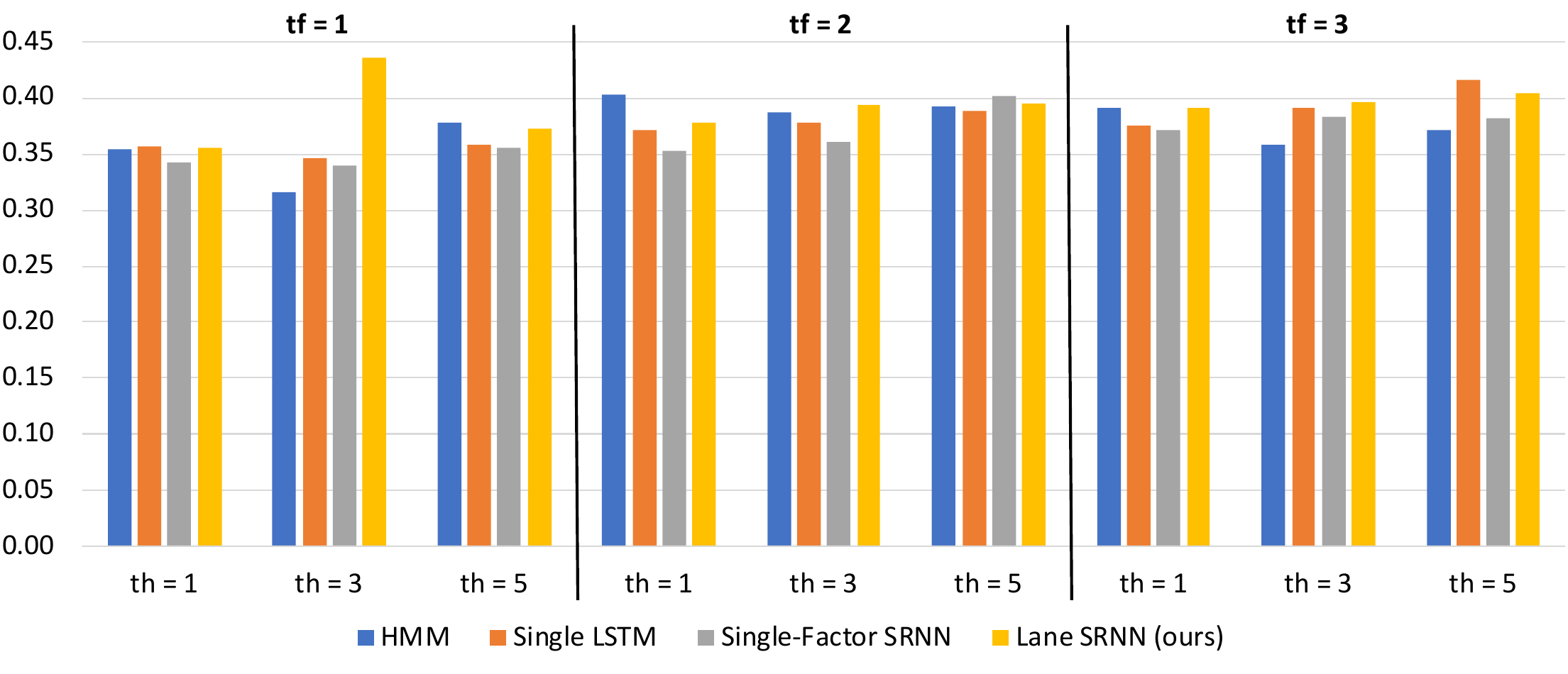}
	\caption{Balanced accuracies for each setting of time history $t_h$ and
		future prediction horizon $t_f$ for our lane SRNN method compared against baseline lane change prediction models.}
	\label{fig:all-ba} 
\end{figure}
From our time horizon experiments, we see a consistent trend in the performance 
	of our Lane SRNN, as well as the RNN baselines, against the classical HMM 
	method with respect to balanced accuracies shown in Fig \ref{fig:all-ba}.
As both the future prediction horizon and time history increases, the RNN-based
	methods increasingly do better than HMMs.
We see the largest performance different for the case of $t_f = 1$ and 
$t_h = 3$ where our lane SRNN outperforms the HMM by 12\% in balanced accuracy. 
In the extreme case of predicting lane changes 3 seconds into the future given 
	5 seconds of history, RNN-based models outperform HMMs by as much as 4\%.
HMMs are limited in their ability to capture intricate temporal lane change 
	models with only multivariate Gaussian emissions and log-linear transition
	probabilities.
Conversely, we see that our RNN-based methods are able to learn richer temporal
	models over longer sequences of time history that lead to thier observed
	higher performance.

\subsection{Comparison of Lane SRNN to RNN Baselines}
From the balanced accuracy results displayed in Fig. \ref{fig:all-ba}, we see
	that our proposed lane SRNN has consistently high performance across all 
	time horizon settings.
In most of the nine time horizon settings, our lane SRNN outperforms both the 
	single LSTM and single-factor SRNN with the highest performance at $t_f = 1$
	and $t_h = 3$.
When compared to the single-factor SRNN specifically, we see that our lane SRNN
	in eight out of our nine time horizon settings.
This points to the merits of the novel three-lane structure within	
 our model over a simpler SRNN model that does not take the structure of the 
 target vehicle's context into account.
Similarly, we see that our SRNN model matches and outperforms the single LSTM 
	model in eight of the nine time horizon cases.
This further shows the benefits of using a high level, interpretable model 
	realized using a composition of RNN units in our lane SRNN over the opaque,
	less transparent single LSTM.
	
While our lane SRNN outperforms the both the single LSTM and single-factor SRNN
	for most cases, there are two independent cases where the either the single
	LSTM or single-factor SRNN perform slightly better than our proposed method
	in terms of balanced accuracies.
These can be seen in Fig. \ref{fig:all-ba} for time horizon settings of two and 
	three second future predictions both given five seconds of time history.
These cases indicate that both the single LSTM and single-factor SRNN can have
	sporadic performance spikes.
We hypothesize that this may be due to the skewed nature of highway driving 
	data; however, we leave the analysis into these two unique cases
	for future work.

Even with these two failure cases, we note that our lane SRNN model demonstrates
	consistently high performance for longer prediction horizons given longer 
	time histories.
In these cases, the consistency of our lane SRNN's performance is more 
	important than one-off, sporadic jumps since predicting lane changes farther
	out into the future given only past observations requires more reliable 
	temporal modeling.
The results of our experiments show that our lane SRNN provides this 
	reliability as opposed to the other RNN baselines.